%% file: main.tex
\pdfoutput=1
\documentclass{article}
\usepackage{iclr2027_conference}

\input{preamble}

\title{Segment Where You Look:\\
       Attention as the Interface Between Language Models and Segmenters}

\iclrfinalcopy
\def\arxivpreprint{}

\author{%
Eslam Abdelrahman$^{1}$\quad
Hamza Omar Alkaf$^{1}$\quad
Yuheng Li$^{2}$\quad
Siddharth Srinivasan Iyer$^{2}$ \\
\bfseries
Nanxuan Zhao$^{2}$\quad
Eli Shechtman$^{2}$\quad
Krishna Kumar Singh$^{2}$\quad
Mohamed Elhoseiny$^{1}$ \\[3pt]
$^{1}$King Abdullah University of Science and Technology (KAUST) \quad
$^{2}$Adobe Research
}

\begin{document}

\maketitle
\ifdefined\arxivpreprint \lhead{}\fi

\input{floats/fig_teaser}

\input{sections/0_abstract/abstract}
\input{sections/1_intro/intro}
\input{sections/2_related_work/related_work}

\input{sections/3_methodolgy/method}
\input{sections/4_exp/experiments}
\input{sections/5_conclusion/conclusion}

\input{sections/6_statements/statements}

\bibliography{refs/related_work}
\bibliographystyle{iclr2027_conference}

\input{sections/6_appendix/appendix}

\end{document}

%% file: preamble.tex
\usepackage{times}
\usepackage[utf8]{inputenc}
\usepackage[T1]{fontenc}
\usepackage{microtype}

\usepackage{amsmath,amssymb,bm}
\usepackage{graphicx}
\usepackage{booktabs}
\usepackage{multirow}
\usepackage{makecell}   %
\usepackage{xcolor}
\usepackage{colortbl}   %
\definecolor{oursblue}{HTML}{DCE6F6}
\definecolor{markgreen}{HTML}{1E7A3C}
\definecolor{markred}{HTML}{B3261E}
\usepackage{pifont}    %
\usepackage{xspace}
\usepackage[font=small,labelfont=bf,skip=4pt,belowskip=0pt]{caption}
\usepackage[font=footnotesize]{subcaption}
\usepackage{hyperref}
\usepackage{url}

\makeatletter
\renewcommand{\paragraph}{\@startsection{paragraph}{4}{\z@}%
  {-0.01ex \@plus -0.01ex \@minus -0.01ex}%
  {-0.6em}%
  {\normalfont\normalsize\bfseries}}
\makeatother

\graphicspath{{figures/}}

\newcommand{\ours}{\texorpdfstring{\mbox{\textsc{AttnSeg}}}{AttnSeg}\xspace}
\newcommand{\seg}{\texorpdfstring{\texttt{[SEG]}}{[SEG]}\xspace}

\newcommand{\attmat}{\bm{A}}              %
\newcommand{\headmap}{\bm{a}}             %
\newcommand{\aggmap}{\hat{\bm{a}}}        %
\newcommand{\gtmask}{\bm{m}}              %
\newcommand{\segemb}{\bm{h}_{\textsc{seg}}}
\newcommand{\nlayer}{L}
\newcommand{\nhead}{H}
\newcommand{\npatch}{P}
\newcommand{\Latt}{\texorpdfstring{\mathcal{L}_{\mathrm{att}}}{L_att}}

\newsavebox{\tabboxa}
\newsavebox{\tabboxb}
\newcommand{\tabheightprobe}[2]{%
  \typeout{[tabprobe] #1 h=\the\ht#2\space d=\the\dp#2\space w=\the\wd#2}}

\newcounter{findingctr}
\renewcommand{\thefindingctr}{\arabic{findingctr}}
\newcommand{\finding}[2]{%
  \par\smallskip\refstepcounter{findingctr}\label{#1}%
  \noindent\fbox{%
    \begin{minipage}{\dimexpr\linewidth-2\fboxsep-2\fboxrule\relax}
      \setlength{\parskip}{0pt}%
      \textbf{Finding~\thefindingctr.} #2
    \end{minipage}}%
  \par\smallskip}

%% file: floats/fig_teaser.tex
\begin{center}
\includegraphics[width=\linewidth]{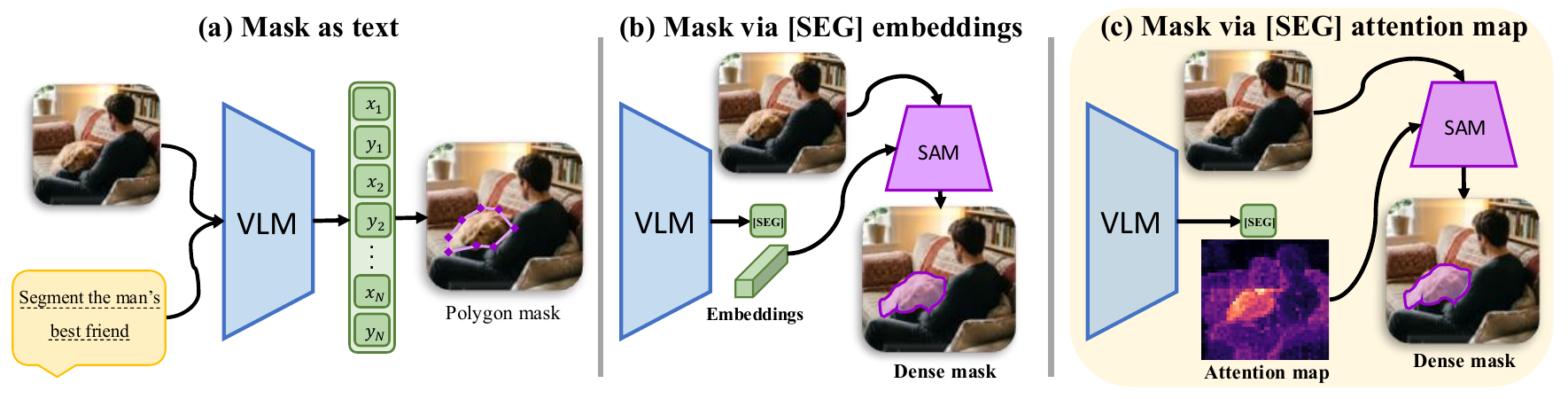}
\captionof{figure}{\textbf{Three ways a vision language model can hand a
segmenter its spatial intent.} (a)~The mask is serialized into the token stream.
(b)~The dominant design: one $d$ dimensional \seg embedding, carrying no spatial
index, prompts SAM. (c)~Ours: the \seg token's attention is already a
distribution on the patch grid, so we forward it instead of re-deriving it.}
\label{fig:teaser}
\end{center}
\vspace{-0.4\baselineskip}

%% file: sections/0_abstract/abstract.tex
\begin{abstract}

Vision-language models (VLMs) that segment from language must pass the location of
the referent to a pixel decoder. The dominant interface compresses it into the
hidden state of a single \seg token, a $d$-dimensional vector with no spatial
index, and leaves the decoder, which never sees the expression, to recover the
referent's extent. We show that the VLM already computes a better signal: the
\seg token's attention over the image tokens, which is defined on the patch grid
by construction. Probing a frozen segmentation VLM, we find that this attention
yields $18$ to $22$ more gIoU than the \seg hidden state when read by the same
probe, that it gains nothing from the image where the hidden state does, and that
it is concentrated in the late layers. Building on these findings, we introduce
\ours, which replaces the \seg embedding with this attention as the decoder's
prompt. A learned, signed combination of the attention heads forms a dense
spatial prompt, and an attention-guidance objective supervises it directly. A
two-stage recipe first trains the VLM to localize without the decoder and then
trains the decoder to refine, within the budget of standard finetuning. Applied to
Sa2VA, \ours improves ReasonSeg by $6.4$ gIoU and GS-Eval by $10.2$ at 4B ($4.2$
and $9.6$ over the same model finetuned on the same data), and by $8.8$ and $11.3$
at 14B. The gains reach $13$ to $18$ gIoU on object parts, multiple objects and
single objects, where the referent must be singled out, and are negligible on
stuff regions, which the decoder already recovers unaided. Code and models are
available at
\href{https://eslambakr.github.io/attseg_project_page/}{\texttt{eslambakr.github.io/attseg\_project\_page}}.

\end{abstract}

%% file: sections/1_intro/intro.tex
\section{Introduction}
\label{sec:intro}

Segmenting an object from a natural-language expression requires deciding
\emph{which} region the expression refers to and \emph{where} exactly its boundary
lies. Promptable segmenters such as SAM \citep{kirillov2023sam}, HQ-SAM
\citep{ke2023hqsam} and SAM3 \citep{carion2026sam} have made the second the easier
half of the problem: given a prompt that points at the right region, they return a
high-quality mask. The first requires understanding the expression, which is the
role of a vision-language model (VLM). This division of labor motivates the hypothesis at the center of our work: \textbf{the VLM should be the load-bearing
localizer, and the decoder should only refine.} Whether this holds depends on the \emph{interface}, the representation through
which the VLM passes the referent's location to the decoder.

Two interfaces dominate (Figure~\ref{fig:teaser}a,b). The first serializes the
mask into the token stream, as polygon vertices
\citep{liu2023polyformer,wang2023visionllm}, learned codewords
\citep{beyer2024paligemma,wang2025himtok} or per-patch labels
\citep{lan2025text4seg}, so that segmentation becomes next-token prediction.
Geometry must then pass through a coarse token bottleneck, and these methods trail
decoder-based systems by a wide margin (Section~\ref{sec:related}). The second,
introduced by LISA \citep{lai2024lisa} and adopted by most subsequent systems
\citep{rasheed2024glamm,zhang2024omgllava,yuan2025sa2va}, adds a \seg token to
the vocabulary and prompts the decoder with its final hidden state, so that the
boundary is drawn at full resolution by a model built for it.

The \seg interface, however, reduces the entire spatial message to a single
$d$-dimensional vector with no spatial index, and the decoder that receives it
never sees the expression. A vector can point at the referent, but it cannot
specify its granularity: for \emph{the left ear of the dog}, a pointer to the dog
leaves the decoder to guess whether the dog, its head or its ear is meant. The
vector is also supervised only through the mask loss computed after the decoder,
so nothing requires the VLM to carry the localization, and the decoder absorbs as
much of it as it can. The released Sa2VA-4B illustrates the consequence. On
GS-Eval \citep{hu2025groundingsuite}, it reaches $90.8$ gIoU on stuff regions,
which the decoder can recover from almost any prompt, but only $49.8$ on single
objects and $17.8$ on object parts. This raises a question: \emph{does the VLM
already compute a better representation of where the referent lies, and can the
decoder be prompted with it directly?}

\paragraph{The VLM already knows where to look.}
We answer the first part by probing a frozen Sa2VA \citep{yuan2025sa2va}. A
lightweight probe must predict the mask from a single internal signal: either the
\seg token's hidden state, which the dominant interface transmits, or its attention
over the image tokens, which is defined on the patch grid by construction. Read by
the same U-Net, the attention yields $18.2$ and $21.5$ more gIoU than the hidden
state on GS-Eval and ReasonSeg \citep{lai2024lisa}
(Figure~\ref{fig:probes:space}). Supplying the RGB image improves the probe of the
hidden state by $4$ to $7$ gIoU and leaves that of the attention unchanged,
consistent with the attention already encoding the referent's extent at patch
resolution. This localization is, moreover, concentrated in the late layers, while
the early layers carry almost none (Figure~\ref{fig:probes:depth}). To our
knowledge, the two signals have not previously been compared within the same
frozen model.

\paragraph{Segmenting where the model looks.}
Building on these findings, we introduce \ours, which replaces the \seg embedding
with the \seg token's attention as the decoder's prompt
(Figure~\ref{fig:teaser}c). A learned $1\times1$ convolution with signed weights
combines the per-head maps into a single map, emphasizing the informative late
heads and subtracting those that attend to sinks or distractors, and the map enters
the decoder's dense prompt channel without being thresholded or sampled. An
attention-guidance objective supervises this map directly. We adopt a weighted
cosine, which asks for the referent's full extent and penalizes concentrated
misplaced attention most. Training proceeds in two stages within the budget of a
single epoch: the VLM first learns to localize under the attention-guidance
objective, without the decoder, and the decoder is then trained to refine the
frozen map. Prior systems derive sparse point prompts from attention or similarity
maps \citep{lin2024vlsam,han2025decaf,qian2025read}; \ours instead reads the \seg
token's own attention, keeps it dense, and trains the VLM to shape it.

\paragraph{Results.}
Applied to Sa2VA, \ours improves ReasonSeg by $6.4$ gIoU and GS-Eval by $10.2$ at
4B, and by $4.2$ and $9.6$ over the base model finetuned on the same data for the
same budget. At 14B, it improves the two benchmarks by $8.8$ and $11.3$ and reaches
$69.7$ gIoU on ReasonSeg without any ReasonSeg training data. The gains
concentrate where the referent must be singled out: object parts, multiple objects
and single objects gain $13$ to $18$ gIoU, while stuff regions gain under $1.5$.
After training, a $560$-parameter gate that can only reweight the model's own maps
recovers close to four fifths of what a $3.86$M-parameter U-Net recovers from the
same attention, against about half before, which shows that the objective
reorganizes the attention into a directly readable form.

In summary, our contributions are threefold:
\begin{itemize}
\item A probing analysis that compares, within a frozen segmentation VLM, the two
  spatial signals of the \seg token. It shows that the attention carries
  considerably more localization than the hidden state the dominant interface
  transmits, and that this localization is concentrated in the late layers.
\item \ours, which replaces the \seg embedding with a learned, signed combination
  of the \seg token's attention maps as a dense decoder prompt, supervised by an
  attention-guidance objective and trained in two stages that make the VLM the
  localizer and the decoder a refiner, within the budget of standard finetuning.
\item Consistent improvements at two scales on ReasonSeg and GS-Eval, concentrated
  on the splits where the referent must be singled out, with analyses
  showing that the attention-guidance objective accounts for most of the gain and makes the attention directly readable.
\end{itemize}

%% file: sections/2_related_work/related_work.tex
\section{Related Work}
\label{sec:related}

We review the literature by interface; Appendix~\ref{app:related} gives an
extended review.

\paragraph{Segmentation with vision-language models.}
A VLM can emit the mask as text, as polygon vertices
\citep{liu2023polyformer,wang2023visionllm,pramanick2024vistallm,xiao2024florence2},
learned codewords \citep{beyer2024paligemma,wang2025himtok,wang2025alto} or
per-patch labels \citep{lan2025text4seg}. These methods trail decoder-based
systems in pixel accuracy, and even Text4Seg gains $5.7$ cIoU on RefCOCO when a
SAM refiner draws the boundary from its labels \citep{lan2025text4seg}, which
indicates that the boundary is the part a decoder absorbs. LISA
\citep{lai2024lisa} instead prompts SAM with the hidden state of a \seg token, an
interface since extended to grounded conversation, multiple and empty targets,
unified pixel-level tasks and video
\citep{rasheed2024glamm,xia2024gsva,zhang2024psalm,zhang2024omgllava,yan2024visa,yuan2025sa2va}.
SAM4MLLM \citep{chen2024sam4mllm} instead emits the box and points of a SAM
prompt as text; we ask how much of the model's localization the \seg vector
carries (Section~\ref{sec:probes}).

\paragraph{What a model's attention encodes.}
Multimodal models localize referents in their attention without grounding
supervision \citep{cao2025difflmm}, although averaging their heads yields noisy
maps: localization is carried by a few heads \citep{kang2025localizationheads},
much of the remaining mass falls on sink tokens \citep{kang2025attentionsink}, and
raw attention needs aggregation before it can be read
\citep{abnar2020rollout,chefer2021transformer}. Deep-layer attention can even
land on the correct region when the model answers wrongly
\citep{liu2025seeingnotbelieving}. These findings motivate our selective, learned
aggregation (Section~\ref{sec:method:aggr}).

\paragraph{Attention as a segmentation prompt.}
VL-SAM \citep{lin2024vlsam} and DecAF \citep{han2025decaf} sample point prompts
from attention without training, and READ \citep{qian2025read} samples points
from an embedding-space similarity map. \ours differs in that it reads the \seg
token's own attention, the internal signal of the dominant interface; it keeps
the map dense in the decoder's mask channel, preserving the graded confidence that
thresholding or sampling would discard; and, unlike the training-free methods, it
trains the VLM so that this attention localizes the referent.

%% file: sections/3_methodolgy/method.tex
\section{Methodology}
\label{sec:method}

This section derives \ours from an analysis of the model it builds on.
Section~\ref{sec:method:prelim} formalizes language-guided segmentation with a VLM
and a promptable decoder, and states our hypothesis about the interface between
the two. Section~\ref{sec:probes} tests that hypothesis by probing a frozen
segmentation VLM, which reveals two properties of its \seg token: its attention
over the image encodes considerably more localization than its hidden state, and
this localization is concentrated in the late layers, while the early layers carry
almost none. 
Building on these findings, Section~\ref{sec:method:approach} introduces \ours,
which replaces the \seg embedding with the attention, combines the attention
heads through a learned aggregation, and supervises the result with an
attention-guidance objective in a decoupled, two-stage training.

\subsection{Problem Setup and Hypothesis}
\label{sec:method:prelim}

The predominant approach to language-guided segmentation pairs a vision-language
model (VLM), which interprets the referring expression, with a promptable
segmentation decoder such as SAM \citep{kirillov2023sam,ravi2024sam2}, which
produces the segmentation mask (Figure~\ref{fig:teaser}b).
Formally, given an image $\bm{I} \in \mathbb{R}^{\hat{H} \times \hat{W} \times 3}$
and a referring expression $\mathcal{T}$, the VLM $\mathcal{F}$ processes both
jointly, drawing on its language understanding and reasoning to determine which
region of $\bm{I}$ the expression describes, and the decoder $\mathcal{D}$
converts that decision into a mask
$\hat{\bm{Y}} \in [0,1]^{\hat{H} \times \hat{W}}$.
The interface between the two adopted by LISA \citep{lai2024lisa}, GLaMM
\citep{rasheed2024glamm}, Sa2VA \citep{yuan2025sa2va} and most of their successors
is the embedding space: a special \seg token is added to the vocabulary, and the
decoder is prompted with the final hidden state of that token alone,
\begin{equation}
\hat{\bm{Y}} = \mathcal{D}\big(\bm{I}, \segemb\big),
\qquad
\segemb = \mathcal{F}(\bm{I}, \mathcal{T})_{r} \in \mathbb{R}^{d},
\label{eq:seg_interface}
\end{equation}
where $r$ is the position of the \seg token in the sequence and $d$ the hidden
width; the learned projection that maps $\segemb$ into the decoder's prompt space
is omitted for brevity. The decoder never observes $\mathcal{T}$, so everything it
learns about the referring expression must pass through a single $d$-dimensional
vector.

\paragraph{A pointer cannot specify granularity.}
Under Eq.~\eqref{eq:seg_interface}, the VLM's role is to indicate where the
referent lies. A pointer suffices when the referent coincides with a region the
decoder can delineate on its own, such as a large, contiguous stuff region, but
it falls short when the expression specifies which object, or which part of it,
is meant.
For instance, given ``\emph{the left ear of the dog}'', a representation that
points at the dog does not determine whether the intended region is the dog, its head or its ear.
Promptable decoders are ambiguous in precisely this way, which is why SAM predicts
several nested masks for a single point prompt \citep{kirillov2023sam}. Since
$\mathcal{D}$ has no access to $\mathcal{T}$, it cannot resolve the ambiguity from
the expression and must fall back on the granularity its training data made most
likely. We therefore hypothesize that compressing the referent into $\segemb$
discards spatial information the VLM has already computed.

\paragraph{Attention as a spatial alternative.}
The VLM already possesses a mechanism whose output is spatial by construction.
Let $\attmat \in \mathbb{R}^{\nlayer \times \nhead \times S \times S}$ collect the
self-attention weights of one forward pass (Figure~\ref{fig:method}, left), where
$\nlayer$ is the number of layers, $\nhead$ the number of heads per layer and $S$
the length of the token sequence, which comprises the image tokens, the text
tokens and the generated answer.
Let $\mathcal{V}$ denote the positions of the $\npatch = hw$ image tokens, which
tile the image on an $h \times w$ patch grid.
For layer $l$ and head $k$, the row queried by the \seg token and restricted to
$\mathcal{V}$ assigns a weight to every image patch,
\begin{equation}
\headmap^{(l,k)} = \big[\, \attmat^{(l,k)}_{r,v} \,\big]_{v \in \mathcal{V}}
\in \mathbb{R}^{\npatch},
\qquad
\headmap^{(l,k)} \leftarrow
\frac{\npatch \, \headmap^{(l,k)}}{\sum_{p} \headmap^{(l,k)}_{p}} .
\label{eq:extract}
\end{equation}
The row is a softmax over the entire sequence, so the share of its mass that
reaches the image varies with sequence length and over training.
Rescaling each map to spatial mean one removes that variation, so a head is
characterized by where it attends rather than by how much.
Folding every map onto the patch grid and stacking them gives
$\headmap \in \mathbb{R}^{\nlayer\nhead \times h \times w}$. Unlike $\segemb$,
each channel is indexed by image location, so it can delineate an object part as
readily as the whole object. Attention maps of this kind are widely read, both to
interpret what a model attends to
\citep{abnar2020rollout,chefer2021transformer,kang2025localizationheads} and as a
signal for localization and prompting
\citep{cao2025difflmm,lin2024vlsam,han2025decaf}. Our hypothesis is that, for
grounding, they carry more of the VLM's localization than $\segemb$ does.
We test it on a frozen model before designing anything on it.

\subsection{What the Model Already Knows, and Where It Keeps It}
\label{sec:probes}

\input{floats/fig_probes}

To validate our hypothesis, we employ probing, the established practice of
training a lightweight classifier on a network's frozen internal representations
to measure what information they encode
\citep{alain2018probes,belinkov2022probing}. We freeze Sa2VA-4B
\citep{yuan2025sa2va} entirely, discard its decoder, and train only a probe that
predicts the ground-truth mask from a single internal signal. Since the model is
frozen, any accuracy the probe attains reflects information the signal already
carries. Two signals are compared: the stacked attention maps $\headmap$ of
Eq.~\eqref{eq:extract}, and the \seg hidden state at every layer, mapped onto the
patch grid by a learned projection shared by all embedding probes. Because a
sufficiently expressive probe can learn the task itself rather than read it from
the representation \citep{hewitt2019controltasks}, each signal is read at two
capacities (Table~\ref{tab:probes}): a single $3\times3$ convolution
(${\approx}5$K parameters) and a U-Net (${\approx}3.86$M). The shared projection
alone costs ${\approx}672$K parameters, far more than the convolutional attention
probe, so any capacity mismatch favors the embedding. Appendix~\ref{app:probes}
gives the architectures and training details, and Section~\ref{sec:exp:analysis}
repeats the procedure on the trained model.

\paragraph{The attention carries more localization than the hidden state.}
Figure~\ref{fig:probes:space} compares the two signals at both capacities. With
the U-Net, the attention reaches $53.2$ gIoU on GS-Eval \citep{hu2025groundingsuite}
and $52.1$ on ReasonSeg \citep{lai2024lisa}, against $35.0$ and $30.6$ for the
\seg hidden state. With the single convolution, by contrast, the two signals score
within a few points of each other, so the attention's advantage appears only once
the probe can combine its channels nonlinearly. This is consistent with prior
observations that localization is carried by a small number of heads, while much
of the remaining attention mass collects on uninformative sink tokens
\citep{kang2025localizationheads,kang2025attentionsink}: the signal is present,
but it must be selected and combined before it can be read.

\finding{f:space}{Read by the same U-Net, the \seg token's attention yields $18.2$
and $21.5$ more gIoU than its hidden state. The vector the dominant interface
transmits is the weaker of the two spatial representations the model computes.}

\paragraph{The image helps the embedding, not the attention.}
A signal that only indicates where the referent lies leaves its extent, the region
it covers, unspecified, and a probe that also sees the pixels should be able to recover that
extent from the image. A signal that already encodes the extent should gain little
from it.
We therefore give the U-Net the RGB image as an additional input, shown as the
shaded segment of Figure~\ref{fig:probes:space}.
The embedding gains $4.4$ and $7.3$ gIoU, whereas the attention changes by $-0.3$
and $+0.2$.

\finding{f:structure}{The image improves the probe of the \seg hidden state by $4$
to $7$ gIoU and leaves the probe of its attention unchanged, consistent with the
attention already encoding the referent's extent at patch resolution, which the
hidden state lacks and must recover from the pixels.}

\paragraph{The signal occupies a narrow late band.}
Having established that the attention is the stronger signal, we ask where in the
network it resides. The answer determines how the maps should be combined, since a
uniform average over layers and heads is appropriate only if localization is
spread evenly across depth. We therefore train a separate probe on each sliding
window of five consecutive layers (Figure~\ref{fig:probes:depth}). Localization
emerges at the same depth for both capacities, which identifies the transition as
a property of the model rather than of the probe. A uniform average would
therefore spend most of its weight on layers that carry no localization; together
with the concentration of localization in a few heads, this motivates the learned
aggregation of Section~\ref{sec:method:aggr}.

\finding{f:depth}{Localization is concentrated in the late layers: roughly the
first $40\%$ of the layers carry almost none, and a single late window of five
layers matches a probe given every layer.}

\subsection{\ours: Segmenting Where the Model Looks}
\label{sec:method:approach}
\label{sec:method:extract}
\label{sec:method:aggr}
\label{sec:method:loss}
\label{sec:method:stages}
\label{sec:method:impl}

\input{floats/fig_method}

\paragraph{Attention as the spatial interface.}
Motivated by these findings, \ours replaces the \seg embedding with the \seg
token's attention as the interface between the VLM and the decoder. Instead of
compressing the referent into $\segemb$, it combines the per-head maps of
Eq.~\eqref{eq:extract} into one map $\aggmap$ and prompts the decoder with it alone:
\begin{equation}
\hat{\bm{Y}} = \mathcal{D}\big(\bm{I}, \aggmap\big),
\qquad
\aggmap = \sum_{l=1}^{\nlayer} \sum_{k=1}^{\nhead} w_{l,k}\, \headmap^{(l,k)}
\in \mathbb{R}^{h \times w},
\qquad
w_{l,k}^{(0)} = \tfrac{1}{\nlayer\nhead} .
\label{eq:aggr}
\end{equation}
Compared with Eq.~\eqref{eq:seg_interface}, the decoder now receives a map defined on the image grid rather than a $d$-dimensional vector. 
The map enters through the decoder's dense mask-prompt channel and it is never thresholded or sampled, so the decoder receives its graded confidence rather than a binarized mask or a handful of points (Appendix~\ref{app:method}). 
The weights $w$ amount to a $1\times1$ convolution over the $\nlayer\nhead$ maps and add only $\nlayer\nhead$ parameters. Their form follows from Section~\ref{sec:probes}: localization is concentrated in a few late heads, so the combination must be selective, and it is learned rather than fixed to the band we measured, which would otherwise have to be located anew for every backbone. Initialized at the uniform average, the combination starts from the plain head mean and departs from it only where training finds a better weighting. 
The weights are also unconstrained in sign: a softmax gate, whose weights lie on the simplex, can only suppress a misleading head, whereas a negative weight can subtract it. 
Appendix~\ref{app:method} gives a second reason to prefer the convolution.

\paragraph{Decoupled two-stage training.}
We decouple the training of the two components, so that the VLM carries the
localization, at the granularity the expression demands, and the decoder only
refines it. In stage one, the VLM and the aggregation weights $w$ are finetuned
with an attention-guidance objective $\Latt$ on $\aggmap$, together with the
standard next-token prediction loss that keeps the model emitting the \seg token.
The decoder takes no part, so no mask loss reaches the map, and $\Latt$ alone
determines what it localizes. In stage two, the VLM and $w$ are frozen and only
the decoder is trained to read the resulting map, which confines it to refining
the localization the VLM provides. Since the map reaches the decoder without
thresholding or sampling, the two stages could be merged into one joint run, and
doing so is how we measured that the decoder's gradient into $\aggmap$ is roughly
$49$ times the magnitude of that of $\Latt$ and roughly orthogonal to it: under
joint training, the map would be shaped mostly by the decoder.
Appendix~\ref{app:method} details what each stage trains and freezes.

\paragraph{Attention-guidance objective.}
Finding~\ref{f:structure} indicates that the attention can encode the referent's extent and not merely its location. 
We train the VLM to strengthen this capability, so that the VLM, not the decoder, resolves which region is meant and at what granularity. 
The map should therefore delineate the referent as fully as a map at patch resolution allows, while recovering the boundary at full resolution remains the decoder's task. 
We supervise $\aggmap$ against the ground-truth mask, area-pooled onto the patch grid as $\gtmask \in \{0,1\}^{\npatch}$ (Figure~\ref{fig:method}, right). 
Since only where the map attends carries information, and not how much attention reaches the image (Eq.~\eqref{eq:extract}), any such objective must be invariant to the scale of $\aggmap$.

\paragraph{Choosing the objective.}
We consider three objectives that are invariant to scale but differ in which parts
of the map they correct. A \emph{peak ratio} (Eq.~\eqref{eq:peak},
Appendix~\ref{app:method}) compares the strongest responses inside and outside the
target. It corrects only those few patches, so it teaches the map where to point
but not how far the referent extends, which reduces the attention to the pointer
that Section~\ref{sec:method:prelim} identified as the limitation of $\segemb$. We
adopt instead a \emph{weighted cosine},
\begin{equation}
\ell^{\cos}(\aggmap, \gtmask) =
1 - \frac{\langle \aggmap, \gtmask \rangle_{\bm{c}}}
         {\lVert \aggmap \rVert_{\bm{c}} \, \lVert \gtmask \rVert_{\bm{c}}},
\qquad
\langle \bm{x}, \bm{y} \rangle_{\bm{c}} = \sum_{p} c_p \, x_p \, y_p,
\qquad
c_p =
\begin{cases}
\lambda_{\mathrm{fg}} & \text{if } \gtmask_p = 1,\\
\lambda_{\mathrm{bg}} & \text{otherwise,}
\end{cases}
\label{eq:cos}
\end{equation}
where $\lambda_{\mathrm{fg}} = 3$ and $\lambda_{\mathrm{bg}} = 1$ up-weight the
foreground. Its natural alternative is a \emph{soft Dice} loss on the min-max
normalized map $\tilde{\bm{a}}$,
$1 - 2\langle \tilde{\bm{a}}, \gtmask \rangle_{\bm{c}} /
(\langle \tilde{\bm{a}}, \bm{1} \rangle_{\bm{c}} + \langle \gtmask, \bm{1} \rangle_{\bm{c}})$.
Both correct every patch and are minimized when the map is uniform over the target
and flat elsewhere, so both ask for the referent's full extent. They differ in how
they penalize attention placed off the target. Dice penalizes its total mass,
$\sum_{p:\,\gtmask_p = 0} \tilde{a}_p$, however that mass is distributed, so its
gradient with respect to $\tilde{\bm{a}}$ is the same for every background patch.
The cosine penalizes it through the norm $\lVert \aggmap \rVert_{\bm{c}}$, which
grows with the square of each response, so its gradient on a background patch is
proportional to the response there,
$\partial \ell^{\cos} / \partial \aggmap_p
= (1 - \ell^{\cos}) \, c_p \, \aggmap_p / \lVert \aggmap \rVert_{\bm{c}}^{2}$
for $\gtmask_p = 0$. For the same amount of misplaced attention, a concentrated
response, such as one locked onto a distractor, therefore costs more than diffuse
background activity, so the cosine spreads residual attention out rather than
concentrating it. A concentrated response is precisely the error that misleads
the decoder, which segments where the map responds most strongly. The cosine also
needs no normalization to be scale-invariant, and it applies directly to the
signed values that Eq.~\eqref{eq:aggr} produces. Section~\ref{sec:exp:analysis}
confirms both arguments by measurement (Table~\ref{tab:loss_ablation}).

\paragraph{Scale-aware weighting.}
The targets in the training mixture span orders of magnitude in area, from object
parts that cover a few patches to stuff regions that fill most of the image, and
the map does not matter equally across this range. The decoder never sees the
expression, so it can correct a poor prompt only through its own objectness
prior, which is reliable for large regions and weak for small ones; the map is
therefore decisive for small targets and largely recoverable for large ones. A
fixed foreground weight cannot express this dependence, so we weight each target
$i$, a \seg token with its map $\aggmap_i$ and mask $\gtmask_i$, by its area,
$s_i = \mathrm{clamp}\big(\bar{m}_i^{-\rho}, s_{\min}, s_{\max}\big)$, where
$\bar{m}_i = \frac{1}{\npatch}\sum_p \gtmask_{i,p}$ is the fraction of the grid
the target covers and $\rho$ sets how strongly small targets are favored. The
two weightings compose at different levels: $\lambda_{\mathrm{fg}}$ and
$\lambda_{\mathrm{bg}}$ balance foreground against background patches within a
target, whereas $s_i$ balances targets against one another.
Upweighting small targets, however, amplifies the gradient in the size range where
the objective is least stable: over training it moves some peaks into the target
and displaces others that were already correct, and a displaced peak costs more
than a corrected one gains. We therefore exclude targets whose map is already
correctly localized, through a gate $g_i$ that is $0$ when the peak of $\aggmap_i$
falls inside $\gtmask_i$ and $1$ otherwise. The objective is
\begin{equation}
\Latt = \frac{\sum_i g_i \, s_i \, \ell^{\cos}(\aggmap_i, \gtmask_i)}
             {\sum_i g_i \, s_i},
\label{eq:latt}
\end{equation}
with $\rho = 0.3$ and $[s_{\min}, s_{\max}] = [0.25, 12]$. Normalizing by
$\sum_i g_i s_i$ keeps the scale of the objective, and hence the effective
learning rate, fixed: gating a target redistributes its weight over the targets
that are still mislocalized instead of shrinking the objective.

\paragraph{What the combination learns.}
The centre of Figure~\ref{fig:method} shows the learned weights on a real forward
pass. Heads that attend to the referent or to a part of it receive positive
weight, whereas a diffuse head, a head collapsed onto an attention sink and a head
locked onto a distractor receive negative weight and are subtracted. No
supervision names these categories: $w$ is trained in stage one by $\Latt$ alone.
Across the trained model about $40\%$ of the weights are negative, and $64\%$ of
the positive mass lies in the last third of the layers. The combination thus
recovers the late band of Section~\ref{sec:probes}, and in addition subtracts
late-layer heads that a softmax weighting could only suppress.
Figures~\ref{fig:qual:gseval} and~\ref{fig:qual:reasonseg} in
Appendix~\ref{app:qual} show further maps beside the masks refined from them.

%% file: floats/fig_probes.tex
\begin{figure}[t]
\centering
\begin{subfigure}[b]{0.393\linewidth}
  \centering
  \includegraphics[width=\linewidth]{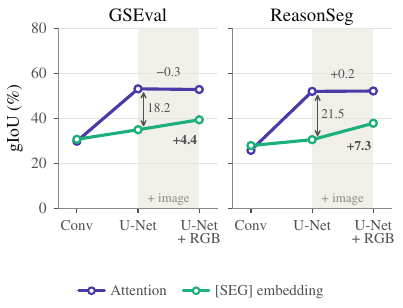}
  \caption{Which space, and what it carries}
  \label{fig:probes:space}
\end{subfigure}
\hfill
\begin{subfigure}[b]{0.595\linewidth}
  \centering
  \includegraphics[width=\linewidth]{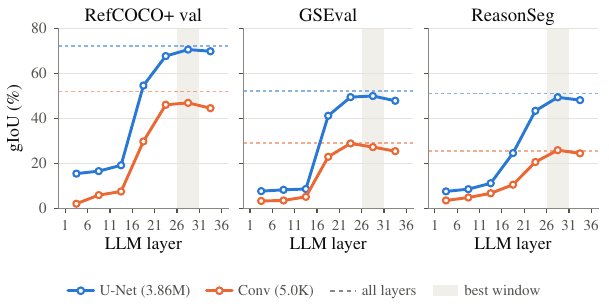}
  \caption{Where in the stack it lives}
  \label{fig:probes:depth}
\end{subfigure}
\caption{\textbf{Which internal signal carries the localization, and where does it
reside?} Probes trained on a frozen Sa2VA-4B. (a)~Read by the same U-Net, the \seg
attention beats the \seg embedding by $18.2$ and $21.5$ gIoU, and adding the RGB
image helps the embedding ($+4.4$, $+7.3$) but not the attention ($-0.3$,
$+0.2$). (b)~Probes restricted to a five-layer window: the early layers carry
almost no localization, and a single late window matches the whole stack
(dashed).}
\label{fig:probes}
\end{figure}

%% file: floats/fig_method.tex
\begin{figure}[t]
\centering
\includegraphics[width=\linewidth]{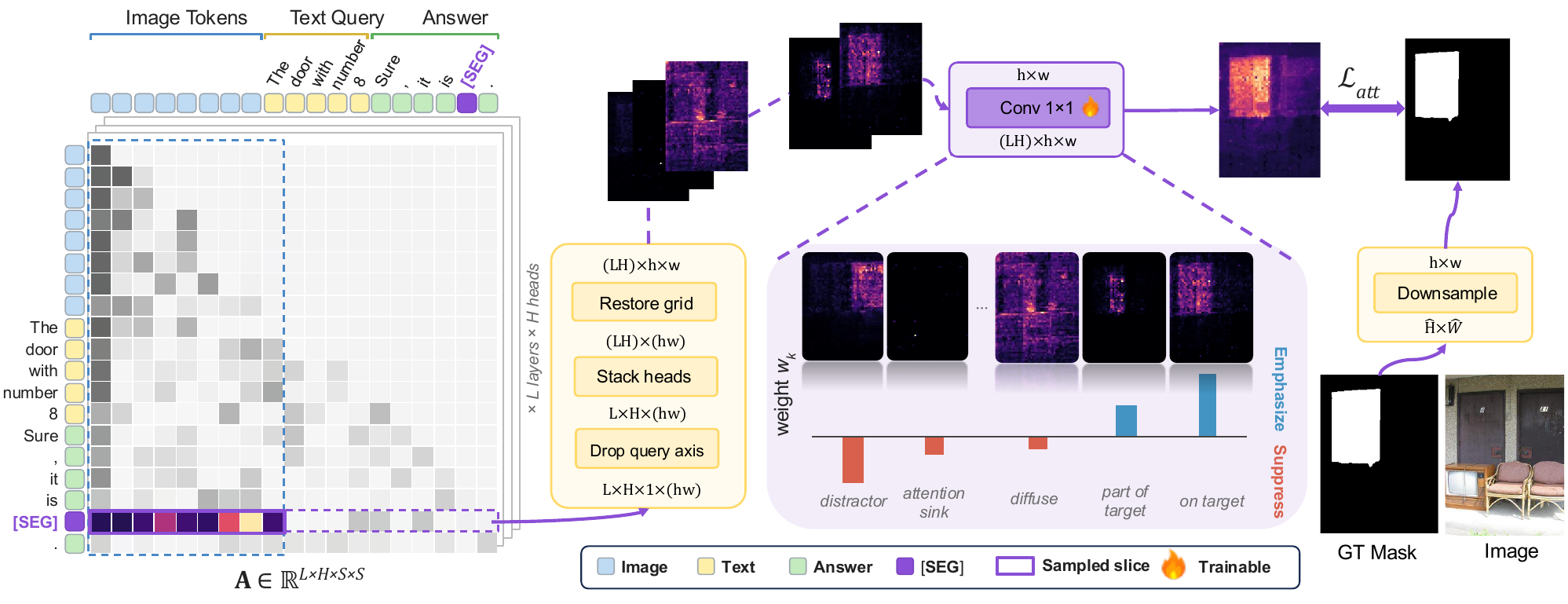}
\caption{\textbf{How \ours reads and shapes the spatial channel.} The \seg
token's query row of $\attmat$, restricted to image columns and folded onto the
patch grid, gives one map per layer and head; a $1\times1$ convolution with
signed weights combines them and $\Latt$ supervises the result.}
\label{fig:method}
\end{figure}

%% file: sections/4_exp/experiments.tex
\section{Experiments}
\label{sec:exp}

We first describe the experimental setup (Section~\ref{sec:exp:impl}), then
compare \ours with its base models and with recent methods
(Section~\ref{sec:exp:main}), and finally analyze where the improvement
originates and how training changes the attention itself
(Section~\ref{sec:exp:analysis}).

\subsection{Experimental Setup}
\label{sec:exp:impl}

\paragraph{Models and training.}
We build \ours on two Sa2VA models \citep{yuan2025sa2va}, whose VLMs are
InternVL2.5-4B and InternVL3-14B, so that the main results are tested at two
scales; the probing and the ablations use the 4B model. Both stages train on the
thirteen-dataset mixture released with Sa2VA, with LoRA of rank $128$ on the
language model and AdamW at a learning rate of $4\times10^{-5}$. Each stage takes
half of a single epoch of this mixture, so \ours uses the same training budget
as finetuning the base model for one epoch on the same data, which is the
baseline of Section~\ref{sec:exp:analysis}.
Reading the attention requires recomputing the \seg rows beside the fused
attention kernel, at the cost of one extra matrix product per layer
(Appendix~\ref{app:limits}). Appendix~\ref{app:exp} lists the mixture and the
remaining settings.

\paragraph{Benchmarks and metric.}
We evaluate on two benchmarks, each challenging for a different reason. The test
split of ReasonSeg \citep{lai2024lisa} names the target by its function or its role in the scene
rather than by its appearance, either as a short phrase or as a full sentence, so
locating it requires reasoning over the image. GS-Eval
\citep{hu2025groundingsuite} instead varies the granularity of the target, over
stuff regions, object parts, multiple instances of one category and single
objects. Neither benchmark's training data is part of our mixture. We report gIoU,
the mean of the per-query IoU, which weighs every query equally; cIoU, which
accumulates intersection and union over the whole set, is dominated by a small
number of very large masks.

\subsection{Main Results}
\label{sec:exp:main}

\input{floats/tab_main}

\paragraph{\ours improves both benchmarks at both scales.}
Table~\ref{tab:main} compares \ours with its base model and with recent
methods. At 4B, \ours raises ReasonSeg from $57.70$ to $64.11$ gIoU and
GS-Eval from $61.91$ to $72.11$, gains of $6.41$ and $10.20$. Part of this gain
comes from finetuning, which Section~\ref{sec:exp:analysis}
isolates: against the base model finetuned on the same mixture at its original
interface for the same epoch, \ours still gains $4.19$ and $9.56$. On ReasonSeg the improvement is
larger for short phrases, from $52.04$ to $61.93$, than for full sentences, from
$59.51$ to $64.80$, consistent with the short form offering less context and
leaving more to localization.

\paragraph{The gain concentrates where the referent must be singled out.}
The GS-Eval splits show where the improvement is earned. Object parts improve from
$17.80$ to $35.29$, multiple objects from $73.33$ to $86.37$ and single objects
from $49.80$ to $62.97$, gains of $17.49$, $13.04$ and $13.17$ gIoU. In each of
these splits the expression must single out its referent among competing
candidates: a part must be separated from the object that contains it, the
instances of one category must be collected while others are excluded, and a
single object must be distinguished from the rest of the scene. Stuff regions are
the exception. The base model already reaches $90.80$ on them, on par with the
strongest methods in the table, and \ours adds $0.40$. A stuff region such as the
sky or a lawn is large and contiguous, so the decoder can recover it from almost
any prompt that falls inside it, which is the regime in which
Section~\ref{sec:method:loss} argued that a poor prompt is recoverable. This
distribution is consistent with our premise. Improvement originating in the
decoder would appear wherever boundaries are hard; here the gain is negligible
where the decoder finds the referent unaided and large where the VLM must
localize it.

\paragraph{The improvement does not diminish with scale.}
An auxiliary objective often benefits a small model most and fades as capacity
grows, on the reading that it supplies something a larger model would acquire by
itself. That is not the case here. On the 14B model, built on the newer
InternVL3, \ours raises ReasonSeg from $60.88$ to $69.71$ and GS-Eval from
$63.05$ to $74.31$, gains of $8.83$ and $11.26$ that exceed those at 4B, and it
does so without any ReasonSeg training data. The split pattern is unchanged:
object parts rise from $26.97$ to $45.01$, multiple objects from $69.66$ to
$85.08$ and single objects from $52.67$ to $66.43$, while stuff moves by $1.40$.
This is consistent with the objective supplying something that neither scale
acquires from the mask loss alone.

\paragraph{A stronger decoder is not a substitute.}
\ours on InternVL2.5-4B outperforms SAMTok \citep{zhou2026samtok}, built on the
more recent Qwen3-VL-4B, by $5.03$ gIoU on ReasonSeg and $10.29$ on GS-Eval.
Strengthening the decoder instead behaves differently. On the same Qwen3-VL-4B
base, replacing SAM2 with SAM3 \citep{carion2026sam} raises ReasonSeg from
$58.67$ to $64.50$, slightly above \ours at 4B, but lowers GS-Eval from $65.41$
to $60.29$, $11.82$ below it. A stronger decoder trades one benchmark for the
other, whereas \ours improves both over its base at both scales, consistent with
the remaining headroom lying in localization rather than in boundary quality.

\subsection{Analysis}
\label{sec:exp:analysis}

\input{floats/tab_loss_ablation}
\input{floats/tab_analysis}
\input{floats/fig_qualitative}

\paragraph{Choice of the attention-guidance objective.}
Table~\ref{tab:loss_ablation} compares the three objectives of
Section~\ref{sec:method:loss} under the same two-stage recipe and area weighting.
The weighted cosine yields the best masks, improving on Dice by $2.20$ and $2.55$
gIoU on ReasonSeg and GS-Eval and on the peak ratio by $3.98$ and $6.12$. Both
margins are significant under a paired bootstrap at $95\%$ confidence, and the
ordering holds at each of four stage-two checkpoints on both benchmarks. The
attention maps explain this ordering along two axes. The peak ratio points best,
placing the strongest patch inside the target for $74.32\%$ and $78.82\%$ of
queries, but specifies almost no extent: its thresholded map reaches $3.26$ and
$4.34$ gIoU, against $49.60$ and $59.09$ for the cosine. This is the pointer of
Section~\ref{sec:method:prelim}, measured directly. Dice specifies extent but
points worst, at $63.89\%$ and $65.65\%$. The cosine ranks second at pointing
and first or second at extent, the only objective with no severe weakness, and
the decoder needs both a correct peak and a specified extent. Its advantage over
Dice does not come from overlap, since on ReasonSeg the Dice map overlaps the
target more, but from where the misplaced attention sits, as
Section~\ref{sec:method:loss} predicts. The cosine leaves more attention off the
target than Dice, $66\%$ against $45\%$ on GS-Eval, yet spreads it out: its
strongest $1\%$ of background patches hold $9\%$ of that mass, against $38\%$
for Dice (Appendix~\ref{app:loss}). Since the decoder segments where the map
responds most strongly, diffuse background is harmless, whereas a concentrated
distractor is not.

\paragraph{Where the improvement originates.}
Table~\ref{tab:ablation} starts from the released 4B checkpoint and adds one
component per row. Finetuning on the training mixture at the original interface
adds $2.22$ gIoU on ReasonSeg and $0.64$ on GS-Eval, and the attention interface
a further $0.26$ and $1.09$. The attention-guidance objective contributes the
remaining $3.93$ and $8.47$ gIoU, $61\%$ and $83\%$ of the total gains of $6.41$
and $10.20$. The interface alone matters much less here than to the probes of
Section~\ref{sec:probes}, where it was worth $18$ to $22$ gIoU, because SAM2 is
far larger than a probe and is trained on the full mixture, so it also learns to
decode the embedding. Its value lies instead in letting $\Latt$ reach the decoder:
the map that $\Latt$ shapes is the map the decoder receives.

\paragraph{Training makes the attention easier to read.}
The benchmark gains could in principle come from the decoder alone learning to
cope with an unchanged map. To separate the two, we repeat the probing of
Section~\ref{sec:probes} on the trained model, again freezing the VLM, and add a
third, much smaller probe: a gate with one weight per attention map ($560$
parameters), which forms a softmax-weighted sum of the maps and can therefore
only select among maps the model has already drawn (Table~\ref{tab:probe_trained}).
Every probe improves, but unevenly. The U-Net gains $1.91$ gIoU on ReasonSeg and
$3.95$ on GS-Eval, whereas the gate gains $18.04$ and $17.86$, from $24.00$ to $42.04$ and from $26.82$ to
$44.68$. The gate thus recovers close to four fifths of what the U-Net recovers,
against about half before training, with nearly seven thousand times fewer
parameters. This is partly by construction, since stage one trains a linear
combination of the same maps, but it shows that $\Latt$ does what it was designed
to do: it adds little that the U-Net could not already extract, and mainly
reorganizes the maps so that a trivial readout can take the localization
directly. Stage two relies on this property, since its decoder trains against a
frozen map. The image also still adds nothing to the attention after training:
given the RGB image as well, the U-Net gains only $0.14$ and $0.30$ gIoU, against
$4.4$ and $7.3$ for the embedding in Figure~\ref{fig:probes:space}. At patch
resolution, the trained map already carries the region it is scored on.

\paragraph{Qualitative results.}
Figure~\ref{fig:qualitative} shows the aggregated map $\aggmap$ beside the mask
the decoder refines from it. Because the map is the decoder's only prompt, the two
can be read against each other directly. Figures~\ref{fig:qual:gseval}
and~\ref{fig:qual:reasonseg} in Appendix~\ref{app:qual} give further examples on
GS-Eval and ReasonSeg.

%% file: floats/tab_main.tex
\begin{table}[t]
\centering
\caption{\textbf{Comparison with prior work}, in gIoU. Each highlighted row is
\ours on the base model directly above it, and bold marks the best value in each
column. Over its base, \ours gains little on stuff and $13$ to $18$ gIoU on
object parts, multiple objects and single objects.}
\label{tab:main}
\setlength{\tabcolsep}{4pt}
\renewcommand{\arraystretch}{1.06}
\resizebox{0.92\linewidth}{!}{%
\begin{tabular}{l ccc ccccc}
\toprule
& \multicolumn{3}{c}{\textbf{ReasonSeg} (gIoU)}
& \multicolumn{5}{c}{\textbf{GS-Eval} (gIoU)} \\
\textbf{Method}
& Sentence & Phrase & Overall
& Stuff & Object parts & Multi obj. & Single obj. & Overall \\
\cmidrule(r){1-1}\cmidrule(lr){2-4}\cmidrule(l){5-9}
LISA-7B                    & 47.49 & 39.08 & 45.47 & 80.34 & 21.25 & 71.95 & 42.62 & 56.34 \\
GLaMM                      & 51.68 & 42.94 & 49.58 & 91.38 & 24.37 & 78.50 & 54.11 & 65.66 \\
Sa2VA-Qwen3-4B-SAM2        & 59.42 & 56.28 & 58.67 & \textbf{91.41} & 22.79 & 76.91 & 54.77 & 65.41 \\
Sa2VA-Qwen3-4B-SAM3        & 65.46 & 61.47 & 64.50 & 89.09 & 17.06 & 70.23 & 48.74 & 60.29 \\
VISA-7B                   & 50.73 & 43.50 & 49.00 & 85.30 & 25.39 & 75.70 & 50.48 & 62.10 \\
SAMTok-Qwen3VL-4B          & 59.41 & 58.03 & 59.08 & 87.64 & 31.01 & 56.77 & 56.28 & 61.82 \\
VideoLISA-3.8B             & 45.59 & 33.61 & 42.72 & 86.99 & 23.40 & 73.54 & 42.52 & 58.70 \\
GSVA-13B                   & 38.73 & 36.75 & 38.26 & 77.34 & 22.84 & 55.46 & 38.30 & 50.58 \\
PixelLM-7B                 & 46.12 & 37.02 & 43.94 & 41.29 & 11.12 & 49.96 & 30.08 & 34.93 \\
\cmidrule(r){1-1}\cmidrule(lr){2-4}\cmidrule(l){5-9}
Sa2VA-InternVL2.5-4B       & 59.51 & 52.04 & 57.70 & 90.80 & 17.80 & 73.33 & 49.80 & 61.91 \\
\rowcolor{oursblue}
\quad + \ours (ours)       & 64.80 & 61.93 & 64.11 & 91.20 & 35.29 & \textbf{86.37} & 62.97 & 72.11 \\
\addlinespace[1pt]
Sa2VA-InternVL3-14B        & 61.44 & 59.11 & 60.88 & 89.45 & 26.97 & 69.66 & 52.67 & 63.05 \\
\rowcolor{oursblue}
\quad + \ours (ours)       & \textbf{69.55} & \textbf{70.23} & \textbf{69.71} & 90.85 & \textbf{45.01} & 85.08 & \textbf{66.43} & \textbf{74.31} \\
\bottomrule
\end{tabular}}
\end{table}

%% file: floats/tab_loss_ablation.tex
\begin{table}[t]
\centering
\caption{\textbf{Why the weighted cosine wins.} Attention-map quality and final
mask gIoU on Sa2VA-InternVL2.5-4B for the three stage-one objectives, each
followed by the same stage two. \emph{Pointing} is the fraction of queries whose
strongest map patch lies inside the target, and \emph{extent} the gIoU of the map
thresholded at $0.5$ after min-max normalization. Each alternative fails on one
axis; the cosine is the only objective with no severe weakness.
Table~\ref{tab:loss_ablation_full} gives per-split results.}
\label{tab:loss_ablation}
\small
\setlength{\tabcolsep}{5pt}
\begin{tabular}{l cc cc cc}
\toprule
& \multicolumn{2}{c}{Map pointing}
& \multicolumn{2}{c}{Map extent}
& \multicolumn{2}{c}{Mask gIoU} \\
Stage-one objective
& ReasonSeg & GS-Eval & ReasonSeg & GS-Eval & ReasonSeg & GS-Eval \\
\cmidrule(r){1-1}\cmidrule(lr){2-3}\cmidrule(lr){4-5}\cmidrule(l){6-7}
Peak ratio             & \textbf{74.32} & \textbf{78.82} & 3.26  & 4.34  & 60.13 & 65.99 \\
Soft Dice              & 63.89 & 65.65 & \textbf{52.64} & 56.03 & 61.91 & 69.56 \\
\rowcolor{oursblue}
Weighted cosine (ours) & 69.24 & 75.77 & 49.60 & \textbf{59.09} & \textbf{64.11} & \textbf{72.11} \\
\bottomrule
\end{tabular}
\end{table}

%% file: floats/tab_analysis.tex
\begin{table}[t]
\centering
\begin{minipage}[t]{0.285\linewidth}
\captionsetup{justification=raggedright,singlelinecheck=false}
\captionof{table}{\textbf{Isolating the source of the gain} on
Sa2VA-InternVL2.5-4B.
}
\label{tab:ablation}
\end{minipage}
\hfill
\begin{minipage}[t]{0.695\linewidth}
\captionof{table}{\textbf{Does $\Latt$ make the attention itself easier to
read?} The probes of Section~\ref{sec:probes}, in gIoU, re-run on the released
Sa2VA-InternVL2.5-4B (baseline) and on \ours built from it. The gain is largest for the smallest probe.}
\label{tab:probe_trained}
\end{minipage}
\par\vspace{2pt}
\begin{minipage}[t]{0.285\linewidth}
\vspace{0pt}
\centering
\setlength{\tabcolsep}{5pt}
\renewcommand{\arraystretch}{1.1}
\savebox{\tabboxa}{%
\begin{tabular}{l cc}
\toprule
Model & ReasonSeg & GS-Eval \\
\cmidrule(r){1-1}\cmidrule(l){2-3}
Baseline                    & 57.70 & 61.91 \\
\quad + Finetuning          & 59.92 & 62.55 \\
\quad + Attention interface & 60.18 & 63.64 \\
\rowcolor{oursblue}
\quad + $\Latt$ (\ours)     & \textbf{64.11} & \textbf{72.11} \\
\bottomrule
\end{tabular}}%
\tabheightprobe{A}{\tabboxa}%
\resizebox{\linewidth}{!}{\usebox{\tabboxa}}
\end{minipage}
\hfill
\begin{minipage}[t]{0.695\linewidth}
\vspace{0pt}
\centering
\setlength{\tabcolsep}{4pt}
\renewcommand{\arraystretch}{1.221}
\savebox{\tabboxb}{%
\begin{tabular}{l cc cc cc cc}
\toprule
& \multicolumn{2}{c}{Gate (560)} & \multicolumn{2}{c}{Conv (5K)}
& \multicolumn{2}{c}{U-Net (3.86M)} & \multicolumn{2}{c}{U-Net + image} \\
Model & ReasonSeg & GS-Eval & ReasonSeg & GS-Eval
& ReasonSeg & GS-Eval & ReasonSeg & GS-Eval \\
\cmidrule(r){1-1}\cmidrule(lr){2-3}\cmidrule(lr){4-5}\cmidrule(lr){6-7}\cmidrule(l){8-9}
Baseline & 24.00 & 26.82 & 25.84 & 29.89 & 52.05 & 53.18 & 52.21 & 52.89 \\
\rowcolor{oursblue}
 + \ours & \textbf{42.04} & \textbf{44.68} & \textbf{31.27} & \textbf{34.67}
                     & \textbf{53.96} & \textbf{57.13} & \textbf{54.10} & \textbf{57.43} \\
\bottomrule
\end{tabular}}%
\tabheightprobe{B}{\tabboxb}%
\resizebox{\linewidth}{!}{\usebox{\tabboxb}}
\end{minipage}
\end{table}

%% file: floats/fig_qualitative.tex
\begin{figure}[t]
\centering
\includegraphics[width=\linewidth]{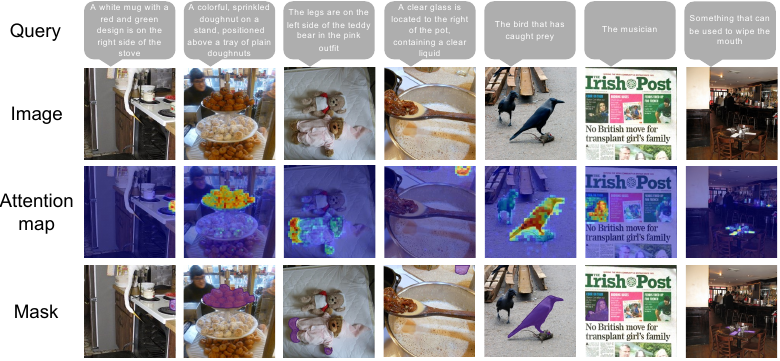}
\caption{\textbf{The attention map is the prediction, not a diagnostic.} For
each query we show the aggregated \seg attention $\aggmap$ and the mask the
decoder refines from it. More examples in
Appendix~\ref{app:qual}.}
\label{fig:qualitative}
\end{figure}

%% file: sections/5_conclusion/conclusion.tex
\section{Conclusion}
\label{sec:conclusion}

We revisited the interface through which a vision-language model passes the
location of its referent to a segmentation decoder. Probing a frozen segmentation
VLM showed that the \seg token's attention carries considerably more localization
than the hidden state the dominant interface transmits, and that this localization
is concentrated in the late layers. \ours builds on these findings: it prompts the
decoder with a learned, signed combination of the \seg token's attention maps in
place of its embedding, supervises that map with an attention-guidance objective,
and trains in two stages that make the VLM the localizer and the decoder a
refiner, within the budget of standard finetuning. At both scales and on both
benchmarks, the gains concentrate on object parts, multiple objects and single
objects, where the referent must be singled out, and are negligible on stuff
regions the decoder recovers unaided. More broadly, our results suggest that
a segmentation VLM already computes where its referent lies, and that the decisive
design choice is to pass that computation to the decoder in the form in which it is
made, and to supervise it there.

%% file: sections/6_statements/statements.tex
\newpage

\section*{AI use statement}

We used generative AI tools as assistants, under the direction of the authors.
For the writing, they helped draft and polish the text: redrafting passages from
author-supplied content and findings, correcting grammar and typography, and
condensing sections. For the code, they helped debug parts of the implementation
and clean it up.

All reported results come from the authors' own runs. No AI system is an author.
We have reviewed all AI-assisted text and code and take full responsibility for
the final content of this work, including every claim, number and artifact.

\section*{Ethics statement}

This work studies the interface between a vision-language model and a promptable
mask decoder. It involves no human subjects and no new data collection: every
dataset and benchmark we use is public and is used as released, under its own
license and evaluation protocol, and both models we build on are publicly
released checkpoints.

Two risks are worth stating plainly. First, our method deliberately moves the
localization burden onto the language model, which means the system's errors
become the language model's errors. It inherits whatever biases the base model
carries in deciding which object a description refers to, and a sharper, more
confident attention map can make a confidently wrong choice of referent harder
for the decoder to correct rather than easier. An improvement in gIoU on the
benchmarks we report is not evidence that the model resolves ambiguous or
contested descriptions correctly, and we would not deploy it in a safety-critical
setting without task-specific validation. Second, a system that segments
arbitrary objects from free-form language lowers the barrier to fine-grained
visual search, and carries the dual-use considerations that come with it.

We have tried to report negative outcomes as carefully as positive ones.
Section~\ref{sec:exp:analysis} reports that the interface substitution on which
the paper is built contributes almost nothing to the end task on its own,
Section~\ref{sec:exp:main} separates the share of the gain that finetuning alone
provides and reports that a baseline with a stronger decoder exceeds \ours at 4B
on ReasonSeg, and Appendix~\ref{app:limits} states the computational cost our
method adds before describing how we reduce it.

\section*{Reproducibility statement}

Section~\ref{sec:method} specifies the method in full: Eq.~\eqref{eq:extract} the
extraction, Eq.~\eqref{eq:aggr} the aggregation, Eq.~\eqref{eq:cos} and
Eq.~\eqref{eq:latt} the objective together with its two refinements, and
Section~\ref{sec:method:stages} the two training stages.
Appendix~\ref{app:method} gives the injection equation, the video case, the
pointing variant we compare against, and the one implementation detail that is
easy to get wrong, namely that freezing the language model does not by itself
freeze the aggregation weights.

Section~\ref{sec:exp:impl} and Appendix~\ref{app:exp} state the complete training
configuration, which is identical at both model scales and in both stages: the
thirteen-dataset mixture released with Sa2VA together with the repetition factor
applied to each, LoRA at rank $128$ with $\alpha=256$ and dropout $0.05$ on the
language model alone, AdamW at $4\times10^{-5}$ with weight decay $0.05$, a
cosine schedule preceded by a $5\%$ linear warmup, gradient clipping at $1.0$,
bfloat16 mixed precision, a maximum sequence length of $8192$, a global batch of
$48$, with each stage taking half of a single epoch. Nothing is tuned per benchmark, so there are no
per-dataset hyperparameters to recover, and every run uses the same seed.
Appendix~\ref{app:probes} gives the probing protocol, including each probe's
architecture, parameter count, input and training budget.

One numerical caveat matters for exact replication. Reading the attention
requires materializing it, which a fused attention kernel does not do, so we keep
the fused kernel for each layer's output and recompute only the \seg query rows
over the full key length with the softmax in full precision
(Appendix~\ref{app:limits}). Those rows are computed exactly as an eager
implementation computes them, but the layer outputs come from the fused path, so
results are not bitwise identical to a run using eager attention throughout.

%% file: sections/6_appendix/appendix.tex
\clearpage
\appendix
\section{Extended Related Work}
\label{app:related}

This appendix extends the review of Section~\ref{sec:related}.

\paragraph{Promptable segmentation and its dense prompt channel.}
SAM \citep{kirillov2023sam} produces a mask at $1024$px from a prompt, and SAM2
\citep{ravi2024sam2} extends it to video with a memory bank that propagates a mask
from a few conditioning frames. Its prompt encoder handles points and boxes
\emph{sparsely}, as positional embeddings concatenated to the decoder's query
tokens, and masks \emph{densely}, downscaling them with a small convolutional
stack and adding the result to the image embedding. The dense channel is the least
used of the three: almost every multimodal segmenter leaves it at SAM's learned
\texttt{no\_mask\_embed} default, and where it is used, it carries a coarse prior
to be cleaned up. SAMRefiner \citep{lin2025samrefiner} mines Gaussian-style mask
prompts from noisy masks so that SAM can repair them, and HQ-SAM
\citep{ke2023hqsam} quantifies how much boundary quality remains recoverable
downstream of a prompt. Because the dense prompt is added to the image embedding,
it accepts a soft, full-field spatial distribution that need not be a mask, which
is exactly the form of an attention map; this is why \ours requires no
thresholding.

\paragraph{Masks as coordinate sequences.}
PolyFormer \citep{liu2023polyformer} cast referring segmentation as sequential
polygon generation before large language models entered the field, regressing
floating-point vertices to avoid quantization error. VisionLLM
\citep{wang2023visionllm} carried the formulation into the LLM era, so that
segmentation becomes text prediction, and VistaLLM \citep{pramanick2024vistallm}
samples contour points adaptively where the boundary bends. Florence-2
\citep{xiao2024florence2} adds $1000$ location tokens to the vocabulary and emits
vertices without a mask decoder, and GiT \citep{wang2024git} routes dense
prediction through a single language interface. VisionLLM's own successor
\citep{wu2024visionllmv2}, however, reintroduces task-specific decoders.

\paragraph{Masks as learned discrete tokens.}
PaLI-3 and PaliGemma \citep{chen2023pali3,beyer2024paligemma} extend the tokenizer
with $128$ codewords from a lightweight segmentation VQ-VAE, so that a mask is
four box tokens followed by sixteen codebook indices. HiMTok
\citep{wang2025himtok} makes the vocabulary hierarchical, representing a mask with
up to $32$ coarse-to-fine tokens that, unlike the \seg route, need no access to the
image at de-tokenization time, and ALTo \citep{wang2025alto} adapts the token
budget to object complexity. These tokens are discrete like text but carry an
inductive bias the language model can exploit, since neighboring codewords denote
similar shapes.

\paragraph{Masks as dense patch labels.}
Text4Seg \citep{lan2025text4seg} labels every patch of a coarse grid with a text
descriptor and compresses the string with a row-wise run-length encoding that
shortens it by $74\%$ and speeds up inference roughly threefold; Text4Seg++
\citep{lan2025text4segpp} extends the recipe. With an InternVL2-8B backbone it
reaches $75.4$ average cIoU on the RefCOCO series, against $69.9$ for LISA and
$75.6$ for GLaMM, the strongest evidence that the text interface is not inherently
weak. Adding a SAM-H refiner on top of its $16\times16$ descriptors lifts RefCOCO
val from $73.5$ to $79.2$ cIoU while leaving the localization decision unchanged,
which indicates that the boundary is the part a decoder absorbs.

\paragraph{Why the text interface loses on geometry.}
Four structural reasons account for the gap. The output resolution is the
language model's coarse patch grid. A token cross-entropy penalizes a vertex that
is off by one bin as heavily as one that is off by a hundred, unless the objective
is explicitly aware of numbers. Under contour or run-length encodings, a one-pixel
shift rewrites many tokens, so small spatial errors become large token errors.
Finally, a single mispredicted early vertex distorts the remainder of the contour.

\paragraph{Extensions of the \seg interface.}
Beyond the systems named in Section~\ref{sec:related}, PixelLM
\citep{ren2024pixellm} replaces SAM with a lightweight decoder and LISA++
\citep{yang2023lisapp} adds instance-level output. The foundational papers argued
against serialization on grounds of reasoning ability and training cost; LISA, for
instance, contrasts VisionLLM's $4\times8$ A100-80G GPUs for $50$ epochs with
under three days on eight 3090s. SAM4MLLM \citep{chen2024sam4mllm} takes a middle
route: it emits the box and points of a SAM prompt as text, so the decoder is
prompted without a dedicated segmentation token.

\paragraph{Attention as localization evidence.}
\citet{kang2025localizationheads} explain why attention-based localization has
looked unreliable: averaging heads gives sparse, noisy maps, whereas a few
\emph{localization heads} track the referent consistently. They isolate these
heads with two criteria, an attention sum that keeps heads placing enough of their
attention on the image and a spatial entropy over the connected components of the binarized map
that keeps heads whose mass forms one compact cluster, and rank heads by how often
they are among the ten lowest-entropy heads over $1{,}000$ image-text pairs. That
selection is made offline and then frozen, whereas the aggregation of
Eq.~\eqref{eq:aggr} learns it from $\Latt$ and admits negative weights, which a
frequency ranking cannot express. \citet{liu2025seeingnotbelieving} further find
that deep-layer attention often lands on the correct evidence region even when
the model answers wrongly, which supports passing the map to the decoder directly and
cautions against treating map sharpness as a proxy for task accuracy.

\section{Probing Protocol}
\label{app:probes}

\input{floats/tab_probes}

Table~\ref{tab:probes} lists the three probes for both signals;
Section~\ref{sec:probes} uses the convolution and the U-Net, and
Section~\ref{sec:exp:analysis} adds the gate. For Sa2VA-4B, the
attention input has $560$ channels, one per (layer, head) pair over every layer
except the first (layers $1$ to $35$ of $0$ to $35$) and all $16$ heads, each
rescaled to spatial mean one as in Eq.~\eqref{eq:extract}; the aggregation of
\ours reads all $36$ layers, hence its $576$ weights. The embedding input stacks
the \seg hidden state of every layer, $36 \times 2048$. All three embedding probes
share one projection onto the grid: a per-layer LayerNorm affine ($147{,}456$
parameters) followed by one $\mathrm{Linear}(2048 \to 16\times16)$ shared across
layers ($524{,}544$), reshaped to a $16\times16$ grid and bilinearly resized to the
sample's resolution.

\paragraph{Architectures.}
The gate learns one scalar per (layer, head) pair, normalized by a softmax over all
$560$ at unit temperature, and outputs the weighted sum of the maps, with no
convolution, bias, normalization or nonlinearity. The convolutional probe is a
single $3\times3$ convolution from $560$ channels to one, again without
normalization or nonlinearity. The U-Net has three stages of base width $64$: a
$3\times3$ convolution from $560$ to $64$ channels, an encoder from $64$ to $128$
to $256$ channels with $2\times2$ max pooling, a $256$-channel bottleneck, a
decoder with $2\times2$ transposed convolutions and skip connections back to $128$
and $64$ channels, and a $1\times1$ convolution to one channel. Every block applies
a $3\times3$ convolution, GroupNorm with one group and a ReLU, twice.

\paragraph{Training and evaluation.}
Each probe trains for three epochs on a fixed training subset, with the VLM frozen
and the decoder unused, and is scored at full image resolution. We report the
probes on RefCOCO+ val, which is in domain for the training mixture, and on
GS-Eval and ReasonSeg, which are not; the latter two carry the evidence of
Section~\ref{sec:probes}.

\paragraph{Thresholding the gate.}
The gate outputs a convex combination of non-negative maps, so after min-max
normalization nearly every pixel exceeds $0.5$, and a fixed threshold of $0.5$
would make its mask all foreground by construction. We therefore report the gate
at a swept threshold, whereas every other probe uses $0.5$. This favors the gate,
so the fractions of the U-Net's accuracy it recovers in
Section~\ref{sec:exp:analysis} are upper bounds; both rows
of Table~\ref{tab:probe_trained} follow the same protocol.

\section{Additional Method Details}
\label{app:method}

\paragraph{Why a signed convolution and not a softmax gate.}
Section~\ref{sec:method:aggr} gives the structural reason: a softmax keeps the
weights on the simplex, so a misleading head can be driven toward zero but never
subtracted. There is also a practical reason. With unbounded logits and a
saturating map, nothing prevents the gate from collapsing onto a handful of heads
during training, and the temperature changes only \emph{when} the collapse occurs,
not whether it does. The $1\times1$ convolution of Eq.~\eqref{eq:aggr} has no
softmax, temperature or normalization, so this failure cannot occur.

\paragraph{The peak-ratio objective.}
Let $\pi$ be the map normalized to unit sum and rescaled by its own detached
maximum, and let $\pi_{\mathrm{fg}}$ and $\pi_{\mathrm{bg}}$ be the means of its
$k$ largest values inside and outside $\gtmask$. The peak ratio compared in
Section~\ref{sec:method:loss} is
\begin{equation}
\Latt^{\mathrm{peak}} = -\log \frac{\pi_{\mathrm{fg}} + \epsilon}
       {\pi_{\mathrm{fg}} + \pi_{\mathrm{bg}} + 2\epsilon} .
\label{eq:peak}
\end{equation}
A margin, $\max(0, \tau - (\pi_{\mathrm{fg}} - \pi_{\mathrm{bg}}))$, is the more
common form of such an objective but fails here. Rescaling by the maximum
compresses both quantities toward one, so the achievable gap saturates, and any
fixed $\tau$ below that ceiling is eventually met by every sample, after which the
term emits no gradient. With $\tau = 0.2$, the fraction of samples still producing
a gradient fell from $0.34$ to zero within the first fifth of training. The ratio
of Eq.~\eqref{eq:peak} is positive for any finite gap and decays smoothly as the
map improves, so it keeps pressure on poor maps without dominating once they are
good.

\paragraph{What each stage trains.}
Stage one finetunes the language model with LoRA, together with the
$\nlayer\nhead$ aggregation weights $w$, under $\Latt$ and the language-modeling
loss that keeps the model emitting the \seg token. The decoder is never
constructed, so no mask loss reaches the map. Stage two freezes the language model
entirely, base weights and LoRA adapters alike, together with $w$, so that
$\aggmap$ becomes a fixed function of the input; $\Latt$ is switched off, and only
the mask branch of the decoder is trained.

\paragraph{Injecting the map into the dense prompt channel.}
We min-max normalize $\aggmap$ per sample, as in Section~\ref{sec:method:loss},
resize it to the resolution of the prompt encoder and map it back to logits,
\begin{equation}
\tilde{\headmap} =
\frac{\aggmap - \min \aggmap}{\max \aggmap - \min \aggmap + \epsilon},
\qquad
\bm{M} = \mathrm{logit}\!\left(
  \mathrm{clamp}\big(\mathrm{resize}_{256}(\tilde{\headmap}),
  \epsilon, 1-\epsilon \big)\right),
\label{eq:inject}
\end{equation}
after which $\bm{M}$ is the decoder's only prompt; the \seg embedding $\segemb$ is
not passed to it. Nothing on this path is thresholded or sampled, so it is
differentiable through Eq.~\eqref{eq:inject} and Eq.~\eqref{eq:aggr} into the
language model's queries and keys. Joint training would use that path; the
two-stage recipe does not, since the language model and $w$ are frozen whenever
the mask loss is active (Section~\ref{sec:method:stages}). On video,
Eq.~\eqref{eq:extract} yields one map per object and conditioning frame, each
frame is prompted with its own map, and the memory bank of SAM2 propagates the
result through the clip.

\paragraph{Freezing the aggregation in stage two.}
Freezing the language model does not by itself freeze the map. The weights $w$ of
Eq.~\eqref{eq:aggr} are registered outside the language model, so a freeze that
covers only its parameters leaves $w$ receiving gradient from the decoder. The map
would then keep adapting to the decoder, and stage two would silently become a
slower form of joint training. Both $w$ and the LoRA adapters must therefore be
frozen explicitly.

\section{Additional Experimental Details}
\label{app:exp}

\paragraph{Training mixture.}
Table~\ref{tab:datamix} lists the thirteen datasets of the training mixture and the
repetition factor applied to each. The mixture, the factors and the seed are
identical in both stages and at both model scales, and the finetuned baselines of
Table~\ref{tab:ablation} use the same mixture and budget, so they differ from
\ours only in the interface and the objective.

\input{floats/tab_datamix}

\paragraph{Benchmarks.}
We evaluate on the test split of ReasonSeg \citep{lai2024lisa}, which reports
separately on its two forms of referring expression, short phrases and full
sentences, and on their union, as in Table~\ref{tab:main}. GS-Eval
\citep{hu2025groundingsuite} reports four splits by the granularity of the target:
stuff regions, object parts, multiple instances of one category and single
objects.

\paragraph{Optimization.}
Both stages use AdamW with a learning rate of $4\times10^{-5}$ and weight decay
$0.05$, a cosine schedule preceded by a linear warmup over the first $5\%$ of the
budget, gradient clipping at $1.0$, bfloat16 mixed precision and a maximum sequence
length of $8192$. Training runs on a single node of H100 80\,GB GPUs with a global
batch of $48$, and each stage takes half of a single epoch of the mixture. The
language model is adapted with LoRA of rank $128$, $\alpha = 256$ and dropout
$0.05$, while the vision encoder, the token embedding, the language-modeling head
and the vision-language projector remain frozen throughout. The same settings are
used at both model scales.

\section{Choice of the Attention-Guidance Objective}
\label{app:loss}

This appendix extends the comparison of Table~\ref{tab:loss_ablation}. All three
objectives run the same two-stage recipe with the area weighting of
Eq.~\eqref{eq:latt}, and the three stage-two runs differ only in the stage-one map
they inherit.

\input{floats/tab_loss_ablation_full}

\paragraph{Per-split results.}
Table~\ref{tab:loss_ablation_full} breaks the comparison down by split. The three
objectives tie on stuff, to within $0.3$ gIoU, a split on which even the released
checkpoint already reaches $90.80$ (Table~\ref{tab:main}), and separate on the
other three splits, where the cosine leads by $2.4$ to $10.4$ gIoU. A paired
bootstrap with $10{,}000$ resamples finds the cosine's advantage over both
alternatives significant on both benchmarks at $95\%$ confidence; Dice and the
peak ratio are significantly different on GS-Eval but statistically tied on
ReasonSeg. The ordering of the three objectives is the same at each of four
stage-two checkpoints on both benchmarks.

\input{floats/tab_loss_concentration}

\paragraph{Where the misplaced attention sits.}
Table~\ref{tab:loss_concentration} measures how each objective distributes the
attention it places outside the target, with map and mask resampled to a common
$256\times256$ grid. The cosine leaves more mass off the target than Dice on both
benchmarks, but far less concentrated: the strongest $1\%$ of background patches
hold $8.9$ to $9.3\%$ of its off-target mass against $37$ to $38\%$ for Dice, and
its participation ratio is three to four times higher. The cosine thus redistributes
misplaced attention rather than removing it, which is the behavior
Section~\ref{sec:method:loss} predicts from its quadratic penalty. The peak ratio
also leaves its off-target mass diffuse, but its maps specify almost no extent
(Table~\ref{tab:loss_ablation}).

\section{Limitations: Reading the Attention}
\label{app:limits}

\ours reads the attention weights, which fused attention kernels never
materialize, and an eager implementation that stores them for the backward pass
does not fit at 14B: $48$ layers by $40$ heads by $\mathrm{seq}^2$ amount to
roughly $44$\,GB in bfloat16 at a typical length of $3400$. Since
Eq.~\eqref{eq:extract} needs only the \seg rows and their image columns, we keep
the fused kernel for each layer's output and recompute just those rows, exactly
as the eager implementation would. The stored tensor shrinks by a factor of
$\mathrm{seq}/R$ for $R$ \seg tokens, at the cost of one extra matrix product
per layer that is linear in sequence length.

\section{Additional Qualitative Results}
\label{app:qual}

Figures~\ref{fig:qual:gseval} and~\ref{fig:qual:reasonseg} show further examples
on GS-Eval and ReasonSeg: for each query, the aggregated map $\aggmap$ and the mask
the decoder refines from it.

\input{floats/fig_qual_gseval}
\input{floats/fig_qual_reasonseg}

%% file: floats/tab_probes.tex
\begin{table}[t]
\centering
\caption{\textbf{The probes.} Section~\ref{sec:probes} uses the convolution and
the U-Net, and Section~\ref{sec:exp:analysis} adds the gate. Three probes of
increasing capacity, applied to both signals with the VLM frozen. The gate has no
convolution, bias, nonlinearity or spatial support, so it can only select among
maps the model has already drawn.}
\label{tab:probes}
\small
\setlength{\tabcolsep}{4.5pt}
\begin{tabular}{llrl}
\toprule
\textbf{Probe} & \textbf{Input} & \textbf{Params} & \textbf{What it can do} \\
\cmidrule(r){1-1}\cmidrule(l){2-4}
\multirow{2}{*}{Gate}
  & attention, $(\nlayer\nhead) \times h \times w$ & 560 & one scalar per (layer, head), softmax, weighted sum \\
  & \seg embedding, $\nlayer \times d$ & 672\,036 & one scalar per layer, then the shared projection \\
\addlinespace[1.5pt]
\multirow{2}{*}{Conv}
  & attention & 5\,041 & a single $3\times3$ convolution, no norm, no nonlinearity \\
  & \seg embedding & 672\,325 & the shared projection, then the same $3\times3$ convolution \\
\addlinespace[1.5pt]
\multirow{2}{*}{U-Net}
  & attention & 3.86\,M & three stages, base width 64, skip connections \\
  & \seg embedding & 4.23\,M & the shared projection, then the same U-Net \\
\bottomrule
\end{tabular}
\end{table}

%% file: floats/tab_datamix.tex
\begin{table}[t]
\centering
\caption{\textbf{Training mixture and sampling weights}, the thirteen datasets
released with Sa2VA \citep{yuan2025sa2va}, identical in both stages and at both
model scales. A factor above one oversamples a dataset within the epoch and a
factor below one is a seeded subsample.}
\label{tab:datamix}
\small
\setlength{\tabcolsep}{7pt}
\renewcommand{\arraystretch}{1.08}
\begin{tabular}{llc}
\toprule
\textbf{Task} & \textbf{Dataset} & \textbf{Repetition factor} \\
\cmidrule(r){1-2}\cmidrule(l){3-3}
\multirow{3}{*}{Semantic segmentation}
  & COCO-Stuff \citep{caesar2018coco} & $0.5$ \\
  & ADE20K \citep{zhou2017scene} & $1$ \\
  & Pascal-Part \citep{chen2014detect} & $4$ \\
\addlinespace[2pt]
\multirow{3}{*}{Referring segmentation}
  & RefCOCO \citep{yu2016refcoco} & $2$ \\
  & RefCOCO+ \citep{yu2016refcoco} & $2$ \\
  & RefCOCOg \citep{mao2016generation} & $2$ \\
\addlinespace[2pt]
\multirow{3}{*}{Referring video segmentation}
  & MeViS \citep{ding2023mevis} & $4$ \\
  & ReVOS \citep{yan2024visa} & $10$ \\
  & Ref-YouTube-VOS \citep{seo2020urvos} & $4$ \\
\addlinespace[2pt]
\multirow{4}{*}{Grounded conversation}
  & GranD-f \citep{rasheed2024glamm} & $10$ \\
  & OpenPSG \citep{zhou2024openpsg} & $1$ \\
  & RefCOCOg-GCG \citep{rasheed2024glamm} & $0.5$ \\
  & Flickr-30k Entities \citep{plummer2015flickr30k} & $0.25$ \\
\bottomrule
\end{tabular}
\end{table}

%% file: floats/tab_loss_ablation_full.tex
\begin{table}[t]
\centering
\caption{\textbf{Choice of the attention-guidance objective, per split}, on
Sa2VA-InternVL2.5-4B, in gIoU. Each row runs both training stages with the area
weighting of Eq.~\eqref{eq:latt} and differs in the stage-one objective. Bold
marks the best value per column.}
\label{tab:loss_ablation_full}
\setlength{\tabcolsep}{4pt}
\renewcommand{\arraystretch}{1.06}
\resizebox{0.92\linewidth}{!}{%
\begin{tabular}{l ccc ccccc}
\toprule
& \multicolumn{3}{c}{\textbf{ReasonSeg} (gIoU)}
& \multicolumn{5}{c}{\textbf{GS-Eval} (gIoU)} \\
\textbf{Stage-one objective}
& Sentence & Phrase & Overall
& Stuff & Object parts & Multi obj. & Single obj. & Overall \\
\cmidrule(r){1-1}\cmidrule(lr){2-4}\cmidrule(l){5-9}
Peak ratio      & 60.75 & 58.16 & 60.13
                & \textbf{91.30} & 28.53 & 80.64 & 52.59 & 65.99 \\
Soft Dice       & 63.06 & 58.28 & 61.91
                & 91.01 & 28.41 & 83.94 & 60.08 & 69.56 \\
\rowcolor{oursblue}
Weighted cosine (ours) & \textbf{64.80} & \textbf{61.93} & \textbf{64.11}
                & 91.20 & \textbf{35.29} & \textbf{86.37} & \textbf{62.97}
                & \textbf{72.11} \\
\bottomrule
\end{tabular}}
\end{table}

%% file: floats/tab_loss_concentration.tex
\begin{table}[t]
\centering
\caption{\textbf{Where the misplaced attention sits.} Properties of the attention
mass outside the target for each stage-one objective: its share of the total map
(off-target), the share of it held by the strongest $1\%$ of background patches
(top-1\%), and its normalized participation ratio, which is $1$ when it is spread
evenly and approaches $0$ when it sits on a single patch.}
\label{tab:loss_concentration}
\small
\setlength{\tabcolsep}{5pt}
\begin{tabular}{l ccc ccc}
\toprule
& \multicolumn{3}{c}{ReasonSeg} & \multicolumn{3}{c}{GS-Eval} \\
Stage-one objective
& Off-target & Top-1\% & Particip. & Off-target & Top-1\% & Particip. \\
\cmidrule(r){1-1}\cmidrule(lr){2-4}\cmidrule(l){5-7}
Peak ratio             & 76.19 & 6.28  & 0.81 & 70.14 & 11.29 & 0.70 \\
Soft Dice              & 45.42 & 36.66 & 0.15 & 44.79 & 38.39 & 0.12 \\
\rowcolor{oursblue}
Weighted cosine (ours) & 64.81 & 8.92  & 0.44 & 66.20 & 9.32  & 0.46 \\
\bottomrule
\end{tabular}
\end{table}

%% file: floats/fig_qual_gseval.tex
\begin{figure}[p]
\centering
\includegraphics[width=0.99\linewidth]{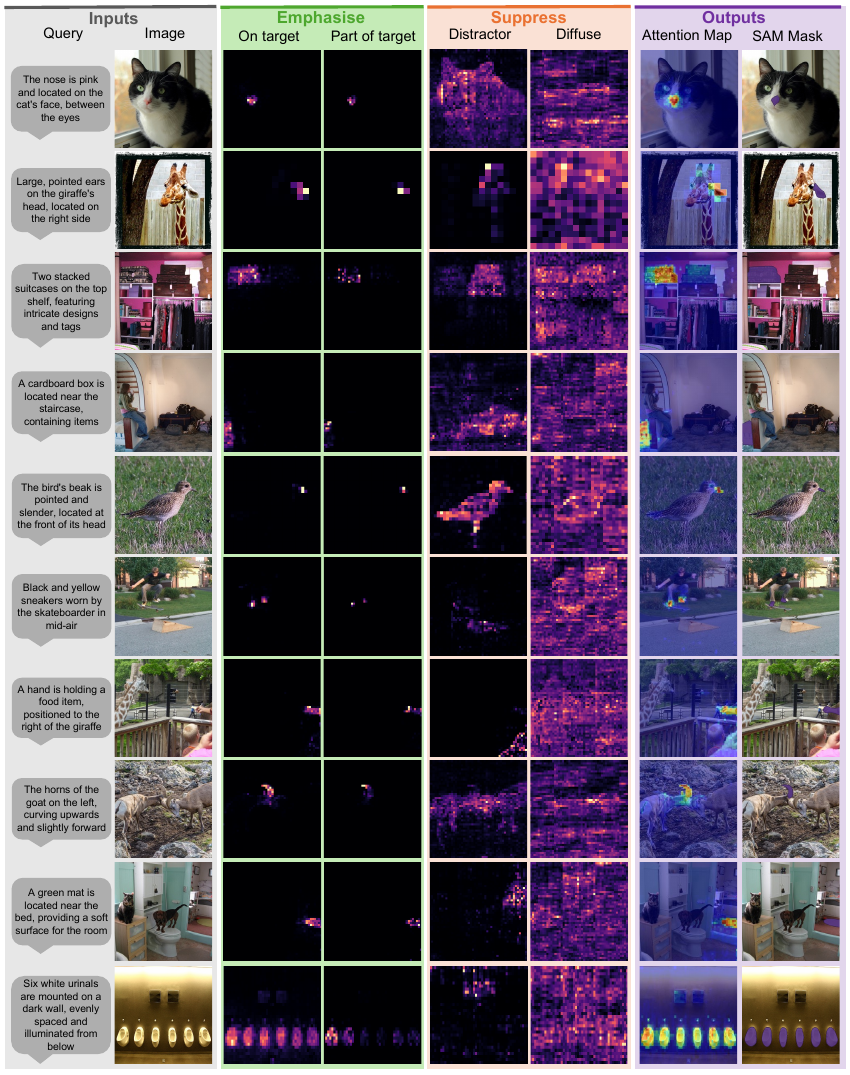}
\caption{\textbf{Additional qualitative results on GS-Eval.} Query, aggregated
\seg attention map, and the refined mask.}
\label{fig:qual:gseval}
\end{figure}

%% file: floats/fig_qual_reasonseg.tex
\begin{figure}[p]
\centering
\includegraphics[width=0.99\linewidth]{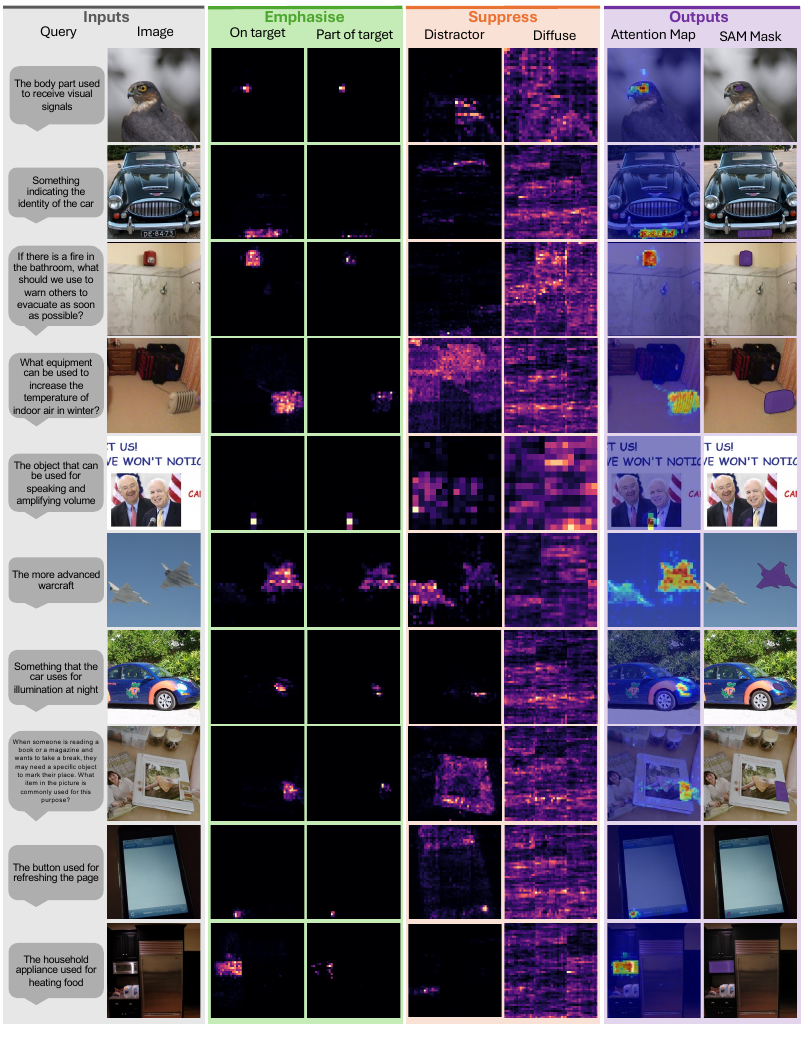}
\caption{\textbf{Additional qualitative results on ReasonSeg.} Query, aggregated
\seg attention map, and the refined mask.}
\label{fig:qual:reasonseg}
\end{figure}